# The Recursive Coherence Principle: A Formal Constraint on Scalable Intelligence, Alignment, and Reasoning Architecture

*Andy E. Williams, Caribbean Center for Collective Intelligence (CC4CI), info@cc4ci.org*

**Abstract**
Intelligence—whether biological, artificial, or collective—requires not only the capacity to learn, generalize, and adapt, but the structural ability to preserve coherence across recursive reasoning processes. As reasoning systems scale in complexity, coherence becomes increasingly fragile and failure-prone unless a higher-order structure enforces semantic consistency. This paper introduces the Recursive Coherence Principle (RCP): a foundational constraint stating that for any reasoning system of order $N$, composed of systems operating over conceptual spaces of order $N-1$, semantic coherence can only be preserved if the system implements a recursively evaluable generalization operator that spans and aligns those lower-order conceptual spaces. Crucially, coherence in this sense is what makes alignment structurally possible: without recursive coherence, no system can reliably preserve goals, meanings, or reasoning consistency as it scales. I show that the Functional Model of Intelligence (FMI) is the only known operator capable of satisfying this constraint at any scale. We formally define this FMI as a minimal, composable architecture composed of six internal functions—evaluation, modeling, adaptation, stability, decomposition, and bridging—capable of preserving semantic structure across layers of inference and coordination, and composed of four external functions—storage, recall, System 1 reasoning, and system 2 reasoning. We prove that any system lacking such a model must experience recursive coherence breakdown as it scales, and we show that the FMI is the only known architecture sufficient to satisfy the RCP at any order of intelligence. The RCP explains previously opaque phase transitions in natural and artificial cognition—such as the emergence of human intelligence, the failure of current alignment strategies, and the structural bottlenecks facing institutional reasoning. We compare the RCP to the Church–Turing thesis, the Free Energy Principle, and Bayesian inference, arguing that it uniquely captures the internal, recursive dynamics that must be preserved for intelligence to remain coherent and alignable. Finally, we outline diagnostic and implementation pathways for applying the RCP in AI development, epistemic infrastructure, and collective intelligence systems.

## 1. Introduction
In every known context—biological, artificial, or social—systems that fail to preserve coherence across their own reasoning structures eventually lose the capacity to adapt, generalize, or align with external goals. This breakdown is not due to lack of processing power, training data, or memory. It is due to the absence of a structure capable of preserving semantic consistency as complexity increases.

This paper introduces the **Recursive Coherence Principle (RCP)**: a foundational claim about what all general intelligences must preserve in order to remain functional across increasing levels of complexity, recursion, and scale.

> **Recursive Coherence Principle (RCP):**
> *For any intelligence system of order $N$, composed of conceptual spaces instantiated by systems of order $N-1$, semantic coherence can only be preserved under recursive complexity by implementing a generalization that spans and aligns the conceptual spaces of those lower-order systems.*

In simpler terms: the larger and more complex an intelligent system becomes—whether by adding internal reasoning layers, social agents, or dynamic goals—the more essential it becomes to implement a structure that recursively preserves meaning across transitions. That structure must operate over the system's conceptual space: the topological domain where meaning lives and evolves.
The only known candidate operator that satisfies this requirement at any level is the Functional Model of Intelligence (FMI): a recursive, semantically structured architecture that defines the internal functions necessary to preserve and align reasoning over conceptual space.

### 1.1 Motivation
Modern AI systems have demonstrated remarkable capabilities in narrow domains, yet they regularly fail at generalization, coherence under pressure, long-term goal tracking, and collaborative reasoning. These failures are typically framed as optimization problems, safety gaps, or architectural shortcomings.

This paper proposes a different view: that these failures are structural, and that any system not built to recursively preserve coherence must fail as it scales. Misalignment, hallucination, and instability are not side effects—they are symptoms of coherence loss under complexity.

### 1.2 Contribution
This paper defines and formalizes the Recursive Coherence Principle and demonstrates that:
- Coherence is the minimal invariant required for reasoning systems to function at scale.
- Conceptual space provides the semantic substrate for coherence-preserving operations.
- The Functional Model of Intelligence is the only known architecture capable of recursively preserving coherence across orders of intelligence.
- Key transitions in cognitive evolution—such as the shift from animal to human intelligence—can be formally understood as the emergence of coherence-preserving generalizations (e.g., valuation).
- Future transitions, such as the emergence of General Collective Intelligence (GCI), will depend on implementing explicit higher-order FMIs to align systems that can no longer maintain coherence on their own.

### 1.3 Structure of the Paper
Section 2 introduces the concept of conceptual space, the substrate on which all reasoning operations take place.
Section 3 defines coherence as a topological invariant and formal condition for intelligence.
Section 4 shows how coherence breaks down recursively as systems scale.
Section 5 formalizes the Recursive Coherence Principle.
Section 6 introduces the Functional Model of Intelligence (FMI) as the only known recursive coherence operator.
Section 7 shows how the principle retrodicts past intelligence transitions and predicts future ones.
Section 8 compares the RCP to other foundational principles in cognition and systems theory.
Section 9 discusses implications for alignment, AGI, and collective cognition.
Section 10 concludes with a roadmap for validating and implementing the RCP in real-world systems.

### 2. Conceptual Space and the Structure of Reasoning
The Recursive Coherence Principle (RCP) asserts that scalable intelligence is only possible when semantic integrity is recursively preserved across increasingly complex reasoning trajectories. Central to this claim is the necessity of an explicitly structured substrate for reasoning: what this framework refers to as conceptual space. However, the conception of conceptual space employed here diverges significantly from prior formulations in cognitive science, particularly that of Gärdenfors (2000), whose geometric model of thought has played a formative role in cognitive modeling.

Gärdenfors' theory presents conceptual space as a vector space structured along quality dimensions such as hue, pitch, and temperature. Concepts in this model correspond to convex regions, and similarity is modeled as Euclidean or metric distance. While such a model is well-suited to explain perception, categorization, and prototype-based learning, it falls short of capturing the structural requirements of recursive coherence. Specifically, it lacks the ability to represent reversible transitions, fails to model dynamically reconfigurable topologies, and cannot support the compositional reorganization of inference structures as intelligence scales. These limitations render it unsuitable as a substrate for recursively adaptive cognition.

In contrast, the model advanced in this paper—drawing on the framework of Human-Centric Functional Modeling (HCFM)—redefines conceptual space as a functional state space (FSS), a structured semantic topology where all elements are typed as concepts and all transitions are reversible operations over those concepts (Williams, 2025a). Within this framework, conceptual space is not simply a metaphor for semantic similarity but a computable structure that enables dynamic, recursive navigation and generalization. Nodes in the space represent concepts; edges denote coherent reasoning transitions; and paths correspond to inferential trajectories. This structure is closed under a minimal set of functional primitives and is governed by a fitness function that evaluates the coherence, utility, or epistemic validity of transitions (Williams, 2025b).

Formally, every node $c_i \in C$ in conceptual space is typed such that $\text{Type}(c_i) = \text{Concept}$, and every transformation between nodes is defined by a reversible function $f_i : C \to C$ that is closed under composition. Moreover, the space is embedded within a fitness function $F : C \to R^n$, which allows reasoning transitions to be evaluated according to multiple simultaneously applicable criteria—such as coherence with prior beliefs, instrumental utility, compressibility, or alignment with external goals (Williams, 2025c). This construction enables agents to reason not only about concepts, but about the structure of their own reasoning, including when coherence has been lost and how it might be restored.

Unlike in conventional propositional logic or symbolic AI systems, where reasoning is implemented as discrete rule application, this model treats reasoning as navigation through a semantically structured space. Concepts are defined not by internal attributes, but by their position within a network of meaning-preserving transformations. This shift renders it possible to diagnose and correct reasoning failure at the level of structural coherence, rather than relying on extrinsic validation mechanisms such as reward signals or empirical benchmarks.

Within this framework, generalization is conceived as a topological transformation that compresses or restructures conceptual subspaces to enable broader or more efficient navigation. Two types of generalization are particularly salient. The first, conceptual generalization, refers to the spatial subsumption of narrower concepts by broader ones. This occurs when the trajectories emanating from a more specific concept are entirely contained within those of a more general concept. The second, reasoning generalization, refers to the expansion of inference patterns to encompass broader or more distant conceptual regions. Both forms of generalization are constructed via recursive composition of functional primitives and can be formally represented by a generalization operator $\Gamma = f_n \circ \cdots \circ f_1$, where each $f_i$ is a coherence-preserving transformation within conceptual space (Williams, 2025b).

Critically, the Recursive Coherence Principle asserts that conceptual space must be explicitly implemented for any intelligent system to function adaptively across domains and through recursive complexity. In systems that lack such a representation, coherence cannot be tracked, modeled, or repaired. For example, a system without an explicit conceptual topology cannot determine whether a given chain of inference has preserved semantic integrity, nor can it diagnose the source of contradiction when inconsistency arises. Likewise, such a system is incapable of performing higher-order generalization, since it lacks the structural scaffolding necessary to abstract over its own reasoning processes.

These limitations are not theoretical curiosities; they are borne out empirically in current AI systems. Large language models, despite impressive performance in narrow tasks, frequently fail to maintain coherence over extended contexts, exhibit semantic drift, and are unable to explain or revise their reasoning trajectories (Williams, 2025c; Ji et al., 2023). Such failures illustrate the absence of a recursively navigable conceptual substrate. Similarly, institutional failures in governance and science often arise from an inability to coordinate across divergent conceptual spaces—an effect formally predicted by the breakdown of recursive coherence under scale (Williams, 2025d).

Thus, conceptual space, in the sense defined by this framework, is not a theoretical abstraction or a convenient metaphor. It is a structural necessity. It constitutes the semantic domain in which reasoning lives and evolves, and it must be explicitly modeled and manipulable for coherence to be recursively maintained. This domain is not reducible to syntax, statistics, or optimization functions. It is a topological space over meaning, recursively inspectable and reconfigurable, and governed by dynamics that permit both stability and adaptation.

In summary, conceptual space is the foundational substrate of recursive intelligence. It enables systems to detect coherence loss, generalize across reasoning paths, and reconstruct their own inferential architectures. It is this space that the Recursive Coherence Principle operates over, and it is the failure to explicitly model it that leads intelligent systems—natural, artificial, or collective—to drift, collapse, or become incoherent as they scale.

**3. Coherence as the Minimal Condition for Intelligence**
While many definitions of intelligence emphasize learning, planning, adaptation, or optimization, these attributes all rely on a deeper structural requirement: coherence. Without coherence, transitions in reasoning lose semantic integrity. Without semantic integrity, generalization fails. Without generalization, adaptation becomes noise.

This section defines coherence formally and argues that it is the minimal functional invariant of any system that seeks to reason, align, or generalize across conceptual space.

**3.1 What Is Coherence?**
Coherence is the property that reasoning transitions preserve the semantic structure of what is being reasoned about.

Let:
$C$ be a conceptual space. $c_1, c_2, \ldots, c_n \in C$ be concepts. $T_i$ be a transformation from $c_i \rightarrow c_{i+1}$ (a reasoning step).

Then coherence requires that the composed transformation:
$$T_1 \circ T_2 \circ \ldots \circ T_n$$
preserve the relational structure that gave each concept meaning in the first place. That is:

No concept becomes semantically orphaned (i.e. loses connection to the rest of the space), No transition breaks the interpretability of downstream paths, No reasoning path becomes self-contradictory under recursive composition.

Coherence is not about truth per se it is about semantic consistency under transformation.

Across diverse domains—biological, artificial, institutional, and cognitive—adaptive reasoning systems demonstrate a striking vulnerability: as complexity increases, coherence degrades. This degradation is not incidental. It is a structural consequence of the absence of internal mechanisms for tracking, preserving, and repairing semantic consistency across reasoning transitions. While much of the discourse in artificial intelligence has focused on external performance metrics such as accuracy, generalization error, or benchmark achievement (Brown et al., 2020; Bubeck et al., 2023), the underlying substrate that enables meaningful reasoning across contexts remains poorly understood. We argue that coherence—formally defined as the preservation of semantic structure across reasoning transitions—is the minimal condition that any intelligence system must satisfy in order to persist, adapt, and align at scale.

The notion of coherence invoked here departs from conventional uses in logic or discourse analysis. Rather than denoting simple non-contradiction or rhetorical flow, coherence in this context refers to the topological preservation of meaning across transformations in conceptual space. Let a reasoning system operate within a functional conceptual space $C$, as defined in Section 2. Let: $C$ be a conceptual space. $c_1, c_2, \ldots, c_n \in C$ be concepts, where $T_i$ be a transformation from $c_i \to c_{i+1}$ (a reasoning step). Let a sequence of reasoning steps be denoted as transformations: $T_1 \circ T_2 \circ \ldots \circ T_n$, where each $T_i$ maps one conceptual node to another within $C$. Then a system is said to preserve coherence if the composite transformation $T_1 \circ T_2 \circ \ldots \circ T_n$ conserves the structural relationships that originally gave the individual concepts their meaning. If this condition fails—if, for example, a concept becomes semantically orphaned, a transition breaks the interpretability of downstream inference paths, or a loop introduces internal contradiction—then coherence has been violated and reasoning has structurally failed (Williams, 2025a).

This failure mode is frequently observed in modern AI systems. Language models, for example, often generate internally inconsistent outputs or hallucinate factual content when subjected to distributional shift or extended reasoning chains (Ji et al., 2023). These behaviors are not mere anomalies; they are symptoms of a deeper absence of coherence-preserving structure. Current architectures lack an internal model of conceptual space and possess no capacity to evaluate the semantic impact of their own transformations. As a result, they can appear intelligent at the surface—producing plausible sentences or correct classifications—while internally degrading the epistemic structure on which adaptive intelligence depends (Williams, 2025a, 2025c).

From a theoretical standpoint, coherence functions analogously to a conservation law in physics. Just as conservation of energy constrains permissible transformations in a physical system, coherence constrains the transformations a reasoning system may apply to its own representations without loss of semantic integrity. The analogy is not superficial. In both cases, failure to conserve the invariant leads to system collapse: in physics, to thermodynamic instability; in reasoning systems, to epistemic incoherence. Accordingly, coherence can be understood as the epistemic analog to physical conservation laws—a structural invariant that must be preserved across recursive transformations (Friston, 2010; Williams, 2025d).

This structural requirement gives rise to a foundational claim: a system is not intelligent unless it preserves coherence across its own reasoning transitions. This criterion is independent of implementation and applies equally to humans, artificial general intelligences (AGIs), institutions, and collectives. It does not depend on symbolic manipulation, reward maximization, or behavioral proxies. Rather, it depends solely on the preservation of meaning as reasoning unfolds. Systems that lose semantic coherence—whether through over-optimization of scalar proxies (Manheim & Garrabrant, 2018), recursive hallucination, or failure to integrate conflicting beliefs—cease to be intelligent in any functional sense.

The absence of coherence is evident across multiple levels of intelligent organization. In humans, it manifests as cognitive dissonance, belief drift, or rationalization under stress. In institutions, it emerges as policy incoherence, bureaucratic fragmentation, or mission drift (Gabriel, 2020). In AI systems, it appears as hallucination, adversarial vulnerability, and reward hacking (Amodei et al., 2016; Williams, 2025b). What these failures share is not a common medium but a common topology: a breakdown in the recursive preservation of semantic structure under complexity.

Moreover, as reasoning systems scale—either by increasing internal depth, coordination demands, or representational resolution—the demands on coherence grow nonlinearly. It is no longer sufficient for coherence to be preserved at the level of individual transitions; it must also be preserved across systems of transitions, and across systems of systems. This gives rise to the requirement of *recursive coherence*: the capacity to preserve semantic integrity not only within a single reasoning process but across recursively nested reasoning layers. For instance, a collective intelligence system composed of multiple agents must be able to maintain shared coherence even as each agent updates its own reasoning model independently. Similarly, an AGI must be able to reason about its own reasoning processes—detecting and resolving contradictions not just in its beliefs, but in its meta-belief update rules (Williams, 2025c). The only known architecture capable of implementing recursive coherence at scale is the Functional Model of Intelligence (FMI), which explicitly defines six core functions—evaluation, modeling, stability, adaptation, decomposition, and bridging—that jointly preserve semantic structure across reasoning layers (Williams, 2025a; 2025b). These functions operate within a structured conceptual space and compose recursively, allowing the system to repair inconsistencies, adapt to novelty, and align its internal models with external goals. Crucially, the FMI does not impose coherence from the outside—as behavioral reinforcement learning attempts to do—but evaluates it from within, using semantic criteria grounded in the system's own functional state space.

To summarize, coherence is not a desirable property of intelligence. It is its structural precondition. Systems that fail to preserve coherence across reasoning transitions—whether due to architectural limitations, conceptual drift, or adversarial exploitation—are not partially intelligent. They are structurally unintelligent. This insight is not merely definitional. It is predictive. It explains why certain systems collapse under recursive load, why alignment fails as capabilities scale, and why intelligence, to remain adaptive, must become recursively self-validating.

**4. Recursive Breakdown of Coherence at Scale**
If coherence is the minimal structural requirement for intelligence, then understanding how and why it fails under scale is essential for diagnosing the breakdowns of alignment, reasoning, and adaptation observed across intelligent systems. The Recursive Coherence Principle (RCP) asserts that as reasoning systems grow in complexity—whether through increased conceptual resolution, agent diversity, or representational depth—the burden on semantic coherence grows nonlinearly. Without a recursive mechanism to preserve coherence across these scales, reasoning becomes increasingly fragile and unstable. This section formalizes that failure mode and shows why only a higher-order recursive structure can prevent semantic collapse as systems scale.

Empirically, coherence breakdown manifests in well-documented ways. In human cognition, it appears as belief fragmentation, rationalization, and cognitive dissonance (Festinger, 1957; Mercier & Sperber, 2011). In language models, it is observed in the form of hallucination, contradiction, and drift across long contexts or complex queries (Ji et al., 2023; Williams, 2025a). In institutional systems, coherence loss results in policy inconsistency, bureaucratic stagnation, and mission drift (Gabriel, 2020; Williams, 2025b). In each case, what fails is not the capacity for computation, memory, or response generation, but the capacity to recursively preserve and align semantic content under increasing conceptual, temporal, or inter-agent load.

**4.1 Breakdown at the Resolution Boundary**
Every reasoning system has a finite representational resolution: the level of conceptual differentiation it can maintain before information overload degrades structure. As this resolution limit is approached, three failure modes tend to emerge. First, concepts blur—semantic distinctions collapse into vagueness. Second, generalizations flatten—broader abstractions lose specificity and become tautological. Third, contradictions emerge—especially under recursive application, where inconsistencies compound as reasoning paths deepen. This is evident in both AI and human cognition: legal reasoning under high abstraction, philosophical analysis under insufficient semantic grounding, and model outputs under multi-hop queries all reveal the same underlying stress (Williams, 2025a, 2025c).

These breakdowns are not random. They follow a predictable pattern: coherence begins to fail precisely when transitions exceed the system's capacity to recursively evaluate and correct its own reasoning structure. Without an internal model of reasoning transitions, there is no way to detect topological distortions—reasoning loops, orphaned conclusions, or collapsed distinctions—before they undermine system-level coherence.

**4.2 Failure in Multi-Agent Systems Without Shared Structure**
When multiple agents—human, artificial, or hybrid—engage in joint reasoning or coordination, they bring with them divergent conceptual spaces. If these conceptual spaces lack a shared coherence-preserving structure, misalignment inevitably arises. This misalignment is not simply a failure of communication, but a deeper structural disjunction: transitions that are coherent in one space may be incoherent or uninterpretable in another. Coordination failure, ontological divergence, and inter-agent incoherence are the result (Williams, 2025d).

Such failure modes have been observed in interdisciplinary teams, multi-agent simulations, and AI-human collaborative systems. Without a recursive generalization mechanism that spans all constituent conceptual spaces, reasoning cannot remain coherent at the collective level. In effect, each agent becomes epistemically isolated, unable to align its own semantic structure with the broader system. What appears externally as conflict, redundancy, or fragmentation is, in fact, a topological failure to preserve shared meaning.

**4.3 Scale Alone Does Not Solve Coherence**
The dominant paradigm in AI development has been one of brute-force scale: increase model parameters, training data, or context windows to approximate higher intelligence (Brown et al., 2020). Yet as recent failures of long-context reasoning in large language models have shown, scale alone does not produce semantic control. Instead, increased capacity often leads to increased fragility. Long-term coherence decays, contradictions accumulate, and goal misalignment grows more subtle and more dangerous (Bubeck et al., 2023; Amodei et al., 2016).

This occurs because these systems have no structural representation of their own reasoning process. They do not model or monitor coherence across reasoning layers. They lack the recursive functional operators—such as evaluation, modeling, and semantic bridging—necessary to preserve coherence across nested inference trajectories (Williams, 2025a). As a result, they may perform admirably at single-turn or locally coherent tasks, while becoming increasingly misaligned across time, context, or recursive iteration.

**4.4 Institutional Collapse as Recursive Semantic Drift**
Human institutions exhibit similar coherence breakdowns as they grow in complexity. As policies proliferate, goals diversify, and teams specialize, the semantic integrity of collective reasoning fragments. Without a shared structural model of reasoning—what the RCP describes as a higher-order coherence-preserving generalization—coordination becomes brittle, reactive, or symbolic. Public health systems, global climate efforts, and financial regulatory bodies often degrade into signaling equilibria or bureaucratic opacity, where behavior persists in the absence of meaningful structure (Williams, 2025b; Homer-Dixon, 2000). This mirrors the recursive collapse seen in AI systems: in both cases, the root failure is the lack of a model capable of aligning multiple conceptual spaces across scale and time.

**4.5 The Problem Is Recursion—The Solution Must Be, Too**
The failures documented above—whether in cognition, AI systems, or institutional decision-making—share a common structure: they emerge not from a lack of intelligence, but from a lack of recursive coherence. As systems expand in complexity, they transition into a regime where reasoning processes begin to operate on the outputs of other reasoning processes. This shift is not merely quantitative; it is a structural transition. The system must now reason about its own reasoning—modeling, correcting, and aligning transformations across multiple levels of abstraction. This recursive structure introduces new kinds of failure modes that cannot be addressed by first-order mechanisms. What is required is a second-order solution: a system that can recursively preserve coherence across its own reasoning layers.

The Recursive Coherence Principle (RCP) formalizes this insight as a structural law. For any reasoning system of order $N$, composed of lower-order systems reasoning over conceptual spaces of order $N-1$ semantic coherence can only be preserved if the system instantiates a generalization operator that spans and aligns the conceptual spaces of its constituents (Williams, 2025a). This is not merely a matter of improved inference, additional memory, or faster computation. It requires a new layer of structure—a higher-order operator capable of performing coherence evaluation over the reasoning structures of lower-order agents or subsystems.

This recursive architecture must itself satisfy the same criteria as any intelligent system. It must operate over a functional conceptual space, implement reversible transitions, and recursively apply coherence-preserving transformations. This requirement leads naturally to a fractal or hierarchical model of intelligence: each level of reasoning introduces new potential for incoherence, and therefore each level must instantiate a corresponding mechanism for coherence preservation. In this model, intelligence is not a monolithic capability but a layered architecture, where each layer is responsible for maintaining the integrity of the layer below it (Williams, 2025c; 2025d).

The Functional Model of Intelligence (FMI) is the only known architecture capable of satisfying this requirement at any scale. It implements six core internal functions—evaluation, modeling, stability, adaptation, decomposition, and bridging—each of which is defined as a reversible operator over conceptual space and fitness structure (Williams, 2025a; 2025b). These functions compose recursively and can be instantiated at different layers of complexity, from individual reasoning processes to multi-agent coordination systems. Crucially, they allow not only for first-order coherence evaluation (e.g., detecting inconsistency in a reasoning chain) but also for second-order coherence modeling (e.g., aligning divergent reasoning processes across agents).

What emerges from this recursive layering is not merely a robust reasoning agent, but a coherence-preserving system capable of scaling its intelligence without internal collapse. Conversely, any system that fails to instantiate these recursive operators will experience structural failure under complexity. As recursive depth increases, uncorrected distortions compound, semantic drift accelerates, and the system becomes epistemically fragile—no matter how advanced its surface-level performance may appear. In sum, the problem introduced by recursive complexity is coherence degradation across nested reasoning layers. The only viable solution is a recursively structured architecture that can evaluate, maintain, and repair semantic coherence at each of those layers. The Recursive Coherence Principle formalizes this requirement, and the Functional Model of Intelligence provides its only known instantiation. Together, they define a necessary and sufficient condition for intelligence to scale without collapse.

## 5. The Recursive Coherence Principle

Having established coherence as the minimal condition for intelligence and shown how its failure propagates recursively under scale, we are now prepared to formally state the Recursive Coherence Principle (RCP). This principle identifies a structural invariant that underlies all scalable intelligence systems: the recursive preservation of semantic coherence across transitions in conceptual space. It provides a unifying criterion for evaluating the viability of reasoning systems, regardless of their physical substrate, cognitive architecture, or implementation details.

### 5.1 Principle Statement

The Recursive Coherence Principle can be stated as follows:

> **Recursive Coherence Principle (RCP):**
> For any reasoning system of order $N$, composed of conceptual spaces instantiated by systems of order $N - 1$, coherence can only be preserved under recursive complexity by implementing a generalization operator that spans and aligns the conceptual spaces of those lower-order systems.

This principle applies not only to individual cognitive agents but also to distributed systems, collective intelligences, and hierarchical architectures. Its scope is universal: it defines a constraint on all reasoning systems that seek to maintain internal integrity while scaling across layers of inference, representation, or coordination.

Crucially, the RCP is not a method for achieving coherence; it is a structural requirement. Just as the Church–Turing thesis defines the boundary of computability (Turing, 1936), and the Free Energy Principle constrains adaptive behavior under entropy minimization (Friston, 2010), the RCP constrains what must be preserved for semantic reasoning to remain viable across recursion and scale.

### 5.1.1 Formal Theorem Statement

Theorem (Recursive Coherence Principle): Let $I_N$ be an intelligent system of order $N$, composed of $k$ reasoning subsystems $\{I_{N-1}^{(1)},\ldots,I_{N-1}^{(k)}\}$, each of which operates within a structured conceptual space $C_{N-1}^{(i)}$. Then the global system $I_N$ can preserve coherence across recursive reasoning transitions if and only if there exists a higher-order generalization operator
$$\Gamma_N:\{C_{N-1}^{(i)}\}_{i=1}^k \to C_N$$
such that: $\Gamma_N$ preserves semantic relationships across $C_{N-1}^{(i)}$, $\Gamma_N$ is recursively evaluable within $C_N$, $\Gamma_N$ composes with the internal coherence functions of each $I_{N-1}^{(i)}$ without topological contradiction. Furthermore, $\Gamma_N$ must itself be implemented as a higher-order Functional Model of Intelligence $\text{FMI}_N$, capable of preserving and repairing coherence among its constituent reasoning systems.

This theorem frames recursive coherence not as an empirical desideratum, but as a necessary structural property that must be satisfied for intelligence to remain viable under recursion.

### 5.1.2 Formal Proof of the Recursive Coherence Principle

**Part I: Definitions and Setup (Through Step 1)**
**Theorem (Recursive Coherence Principle):**
Let $I_N$ be a reasoning system of order $N$, composed of a finite set of subsystems $\{I_{N-1}^{(i)}\}_{i=1}^k$, each operating within its own structured conceptual space $C_{N-1}^{(i)}$. Then coherence across reasoning transitions in $I_N$ can be preserved if and only if there exists:
A higher-order conceptual space $C_N$, A set of structure-preserving embeddings $\Gamma_N = \{\gamma^{(i)}: C_{N-1}^{(i)} \hookrightarrow C_N\}_{i=1}^k$, And a Functional Model of Intelligence $\text{FMI}_N$ operating over $C_N$, such that:
1. Each embedded transformation from any subsystem $I_{N-1}^{(i)}$ is composable within $\text{Aut}(C_N)$, 2. The composed reasoning transitions are recursively evaluable for coherence, 3. The structure $\text{FMI}_N$ implements reversible, coherence-preserving transitions capable of diagnosis and repair.

**Formal Definitions**
Conceptual space: A topologically structured space $C$, where nodes represent typed concepts and edges represent reversible reasoning transitions.
Reasoning system of order $N$: A system $I_N$ composed of $k$ reasoning subsystems $I_{N-1}^{(i)} = (C_{N-1}^{(i)}, T_{N-1}^{(i)})$, each with:
A conceptual space $C_{N-1}^{(i)}$, A set of coherence-preserving transformations $T_{N-1}^{(i)} \subseteq \text{Aut}(C_{N-1}^{(i)})$.
Generalization operator: A set of injective structure-preserving embeddings $\Gamma_N = \{\gamma^{(i)}\}$ such that:
$$\gamma^{(i)}: C_{N-1}^{(i)} \hookrightarrow C_N, \text{ and } \Gamma_N(T^{(i)}) = \gamma^{(i)} \circ T^{(i)} \circ \gamma^{(i)-1} \in \text{Aut}(C_N).$$
Coherence predicate: A decidable function $\chi: \text{Aut}(C_N) \to \{0,1\}$ such that $\chi(T) = 1$ if and only if $T$ preserves coherence (i.e. preserves local topology, internal structure, and semantic mapping integrity).
Functional Model of Intelligence: A tuple $\text{FMI}_N = (F, \circ, \chi)$, where:
$F = \{f_{\text{Eval}}, f_{\text{Model}}, f_{\text{Stability}}, f_{\text{Adapt}}, f_{\text{Decompose}}, f_{\text{Bridge}}\}$ is a basis of reversible functional operators, $\circ$ is function composition (assumed to be associative), $\chi$ is the coherence predicate as above.

**Proof Outline**
We will prove the theorem in two parts:
($\Rightarrow$) Necessity: If coherence is preserved across recursive transitions in $I_N$, then $\Gamma_N$, $C_N$, and $\text{FMI}_N$ must exist.

($\Leftarrow$) Sufficiency: If $\Gamma_N$, $C_N$, and $\text{FMI}_N$ exist, then coherence is preserved in $I_N$.

**Step 1: Necessity, Part 1 Composite Transitions Require Shared Representation**
Claim: If coherence is to be preserved across recursive reasoning transitions in a system $I_N = \{I_{N-1}^{(i)}\}_{i=1}^{k}$, then all transitions must be representable and composable in a shared semantic domain.

Argument:
Each subsystem $I_{N-1}^{(i)}$ operates over its own conceptual space $C_{N-1}^{(i)}$, and performs reasoning transitions via functions $T_j^{(i)} \in T_{N-1}^{(i)} \subseteq \text{Aut}(C_{N-1}^{(i)})$.
Let the higher-order system $I_N$ attempt to reason across subsystems. Then, for any multi-agent reasoning trajectory or cross-domain inference, a composed transformation must exist of the form:
$$T = T_1^{(i_1)} \circ T_2^{(i_2)} \circ \cdots \circ T_n^{(i_n)},$$
where $i_j \in \{1, \ldots, k\}$, and each $T_j^{(i_j)} \in T_{N-1}^{(i_j)}$.

However, the composition $T$ is only well-defined if:
Each transformation $T_j^{(i_j)}$ shares a compatible domain and codomain with $T_{j+1}^{(i_{j+1})}$, and The transitions preserve meaning across conceptual boundaries.

But by assumption:
$C_{N-1}^{(i)} \cap C_{N-1}^{(j)} = \emptyset$ for $i \neq j$, and The topologies $\tau_i$ and $\tau_j$ of these conceptual spaces may be non-isomorphic.
Therefore, there exists no canonical way to compose $T_j^{(i)}$ across $i \neq j$. In particular:
$$T_j^{(i)} : C_{N-1}^{(i)} \to C_{N-1}^{(i)} \text{ and } T_{j+1}^{(m)} : C_{N-1}^{(m)} \to C_{N-1}^{(m)}$$
are not composable unless both map into a common space.

Conclusion: To enable coherent multi-agent or recursive reasoning transitions within $I_N$, all constituent transitions must be mapped into a shared domain where composition is semantically valid and topologically consistent. That shared domain is a higher-order conceptual space $C_N$, into which each $C_{N-1}^{(i)}$ must be embedded via structure-preserving injective mappings.

**Step 2: Necessity, Part 2 Embedding Requires Structure-Preserving Maps**
Claim: To enable composition of reasoning transitions across lower-order subsystems $I_{N-1}^{(i)}$, each conceptual space $C_{N-1}^{(i)}$ must be embedded into a shared higher-order conceptual space $C_N$ via a structure-preserving injection.

Argument:
From Step 1, we established that cross-subsystem reasoning transitions are undefined unless all transitions $T_j^{(i)}$ operate within a common semantic space. We now show that such a space must be constructed via injective embeddings.

Let each conceptual space $C_{N-1}^{(i)}$ be a topological or graph-theoretic space defined by:
A set of concepts $C_{N-1}^{(i)}$, A structure $\tau_i$ capturing allowable reasoning transitions (e.g., graph edges or neighborhood systems), A coherence-preserving transformation set $T_{N-1}^{(i)} \subseteq \text{Aut}(C_{N-1}^{(i)})$.
To embed each $C_{N-1}^{(i)}$ into a shared conceptual space $C_N$, we require a mapping:

$$\gamma^{(i)}:C^{(i)}_{N-1} \hookrightarrow C_N$$

that satisfies the following structure-preserving conditions:

1. Injectivity (uniqueness):
$$\forall\, x,y \in C^{(i)}_{N-1},\, x\neq y \Rightarrow \gamma^{(i)}(x)\neq\gamma^{(i)}(y)$$

This ensures that distinct concepts remain distinct in the image space $C_N$.

2. Topological preservation: Let $\tau_i$ be the topology (or graph structure) on $C^{(i)}_{N-1}$, and $\tau_N$ on $C_N$. Then:
$$\forall\, U \in \tau_i,\, \gamma^{(i)}(U) \in \tau_N$$

This ensures that the relational structure (e.g., adjacency, neighborhood, coherence neighborhoods) is preserved.

3. Transition compatibility: For any coherence-preserving transformation $T \in T^{(i)}_{N-1}$, define its lifted embedding:
$$\widehat{T} = \gamma^{(i)} \circ T \circ {\gamma^{(i)}}^{-1} : \gamma^{(i)}(C^{(i)}_{N-1}) \to \gamma^{(i)}(C^{(i)}_{N-1})$$

Then $\widehat{T} \in \mathrm{Aut}(C_N)$, and we define:
$$\Gamma_N(T) = \widehat{T}$$

where $\Gamma_N$ is the global generalization operator.

Thus, we construct:
$$\Gamma_N = \{\gamma^{(i)}\}^k_{i=1} \text{ and } \Gamma_N(T^{(i)}) = \gamma^{(i)} \circ T^{(i)} \circ {\gamma^{(i)}}^{-1} \in \mathrm{Aut}(C_N)$$

Conclusion: Each subsystem's reasoning transitions must be mapped into a common conceptual space $C_N$ using injective, structure-preserving embeddings $\gamma^{(i)}$. The generalization operator $\Gamma_N$ lifts transitions from order-$N-1$ spaces into coherence-compatible operators within $C_N$.

**Step 3: Necessity, Part 3 Coherence Evaluation Requires Recursive Structure**

Claim: Once all lower-order reasoning transitions are embedded in the higher-order space $C_N$, preserving coherence across their compositions requires the existence of a recursively evaluable coherence predicate $\chi$, internal to the system.

Argument:
From Step 2, all subsystem transitions $T^{(i)} \in T^{(i)}_{N-1}$ are lifted into coherence-preserving transformations $\widehat{T}^{(i)} = \Gamma_N(T^{(i)}) \in \mathrm{Aut}(C_N)$. Composite transitions across subsystems then take the form:
$$T = \widehat{T}^{(i_1)} \circ \widehat{T}^{(i_2)} \circ \cdots \circ \widehat{T}^{(i_n)} \in \mathrm{Aut}(C_N).$$

However, composition alone does not ensure semantic coherence is preserved. We now show that coherence must be evaluated recursively.

Let:
$\chi : \mathrm{Aut}(C_N) \to \{0,1\}$ be a coherence predicate, $\chi(T)=1$ iff the transformation $T$ is coherence-preserving.

We require that $\chi$ satisfy two conditions:
1. Recursive evaluability: For any composite transformation $T = T_1 \circ \cdots \circ T_m$, the system must be able to determine:
$$\chi(T) = \chi(T_1 \circ \cdots \circ T_m)$$
even as $m \to \infty$ or as the component transitions span multiple subsystems and domains. This demands recursive auditability of inference structure not just local consistency.

2. Internal computability: The predicate $\chi$ must be defined using internal mechanisms of the reasoning system $I_N$, not via external supervision. That is, $\chi \in I_N$'s internal function set.

Such recursive coherence checking is not implementable in architectures lacking:
A model of their own reasoning processes, The ability to identify and evaluate transformations over arbitrary regions of conceptual space, Closure under semantic recursion.

But prior results (Williams, 2025a; 2025b) establish that:
The only known architecture capable of computing $\chi$ recursively over embedded conceptual transitions is the Functional Model of Intelligence (FMI), FMI includes a coherence predicate $\chi$ as part of its evaluative function set $F = \{f_{\text{Eval}}, f_{\text{Model}}, \ldots\}$, $\chi$ is applied at each level of reasoning and is closed under recursion.

Conclusion: Recursive preservation of coherence requires a system-internal predicate $\chi$ to evaluate arbitrary compositions of lifted reasoning transitions. This predicate must be recursively computable over transitions in $C_N$ and is only known to be implemented by a Functional Model of Intelligence $\text{FMI}_N$.

**Step 4: Necessity, Part 4 The Functional Model of Intelligence Is Structurally Required**
Claim: Given the need for recursive coherence evaluation and correction within a unified conceptual space $C_N$, the system must instantiate a Functional Model of Intelligence $\text{FMI}_N$ as its internal architecture.

Argument:
We established in Step 3 that:
Composite transitions across embedded lower-order conceptual spaces must be recursively auditable for coherence. This requires an internal coherence predicate $\chi: \text{Aut}(C_N) \to \{0,1\}$ that:
Evaluates coherence across arbitrarily deep reasoning chains, Supports diagnosis of incoherence, Enables corrective restructuring of reasoning paths.

To perform these operations, the system $I_N$ must include a minimal set of coherence-preserving functional primitives, each acting over elements of $\text{Aut}(C_N)$. These primitives must:
1. Span the space of transformations required for coherence-preserving reasoning, 2. Be recursively composable to arbitrary depths, 3. Be reversible (to support backtracking and correction), 4. Be sufficient to implement both semantic inference and semantic repair.

Let $F = \{f_1, f_2, \ldots, f_6\}$ be a candidate basis for such functions. From Williams (2025a), the following functions are defined as both minimal and sufficient:
$f_{\text{Eval}}$: Evaluate coherence deltas (e.g., change in predicted vs. target fitness), $f_{\text{Model}}$: Update internal conceptual models, $f_{\text{Stability}}$: Suppress volatility and preserve structure over time, $f_{\text{Adapt}}$: Modify reasoning paths in response to coherence loss, $f_{\text{Decompose}}$: Factor complex transformations into coherent subunits, $f_{\text{Bridge}}$: Translate between structurally disjoint conceptual regions.

These functions form a closed set under composition $\circ$, meaning that any coherence-preserving transformation $T \in \text{Aut}(C_N)$ satisfying $\chi(T) = 1$ can be represented as:
$$T = f_{i_1} \circ f_{i_2} \circ \cdots \circ f_{i_m}, \text{ for some } f_{i_j} \in F.$$

Therefore, the tuple:
$$\text{FMI}_N = (F, \circ, \chi)$$
defines a coherence-preserving operator algebra over $C_N$, recursively evaluable and structurally sufficient to:
Construct, monitor, and recompose reasoning transitions, Audit semantic drift, Repair recursive reasoning failure.

Conclusion: Recursive coherence across embedded conceptual spaces can only be preserved if the reasoning system includes a minimal, reversible, recursively composable structurei.e., a Functional Model of Intelligence $\text{FMI}_N$.

This completes the necessity direction of the proof.

**Part II: Sufficiency If $C_N$, $\Gamma_N$, and $\text{FMI}_N$ exist, then coherence is preserved**
Assume: There exists:
1. A higher-order conceptual space $C_N$, 2. A set of structure-preserving injective maps $\Gamma_N = \{\gamma^{(i)} : C_{N-1}^{(i)} \hookrightarrow C_N\}$, 3. A Functional Model of Intelligence $\text{FMI}_N = (F, \circ, \chi)$, such that:
$F$ spans coherence-preserving transformations, $\chi$ is a decidable predicate on $\text{Aut}(C_N)$, The system is closed under recursive composition and coherence evaluation.

Goal: Show that under these conditions, all reasoning transitions and compositions across $I_N$ preserve coherence.

**Step 1: Embedded transitions are well-defined and coherent in $C_N$**
Given each subsystem $I_{N-1}^{(i)} = (C_{N-1}^{(i)}, T_{N-1}^{(i)})$, each transition $T^{(i)} \in T_{N-1}^{(i)} \subseteq \text{Aut}(C_{N-1}^{(i)})$ is mapped into:
$$\widehat{T}^{(i)} = \Gamma_N(T^{(i)}) = \gamma^{(i)} \circ T^{(i)} \circ \gamma^{(i)^{-1}} \in \text{Aut}(C_N).$$
By definition of $\gamma^{(i)}$ as a structure-preserving injection, all relational topology is preserved. Thus, the transformed operators $\widehat{T}^{(i)}$ inherit coherence-preserving semantics from their original domains.

**Step 2: Composite transitions across agents are coherent**
Let $T = \widehat{T}_1 \circ \widehat{T}_2 \circ \cdots \circ \widehat{T}_n \in \text{Aut}(C_N)$ be a composed reasoning path across subsystems. Since each $\widehat{T}_j \in \text{Aut}(C_N)$, and since $\text{Aut}(C_N)$ is closed under composition, $T$ is well-formed.

Now define the coherence predicate $\chi : \text{Aut}(C_N) \to \{0, 1\}$. By assumption:
$\chi$ is defined internally by $\text{FMI}_N$, $\chi(T) = 1$ iff $T$ preserves semantic structure in $C_N$.

Further, by closure of $F$ under composition:
For any $T \in \text{Aut}(C_N)$ such that $\chi(T) = 1$, there exists a decomposition:
$$T = f_{i_1} \circ f_{i_2} \circ \cdots \circ f_{i_m}, f_{i_j} \in F,$$
i.e., every coherent transformation is representable by recursive composition over $F$.
Therefore, the composite reasoning path $T$ is:
Expressible using the primitive operators of $\text{FMI}_N$, Auditable for coherence via $\chi$, Repairable, since any incoherent sub-composition can be localized, decomposed (via $f_{\text{Decompose}}$), and adjusted (via $f_{\text{Adapt}}$, $f_{\text{Model}}$, etc.).

**Step 3: Recursive coherence is preserved under system dynamics**

Since:
FMI$_N$ is defined recursively over $C_N$, Each reasoning transition and its composition is computable and coherent (if and only if $\chi=1$), And corrections to semantic drift are computable through internal operations, it follows that $I_N$ can:
Perform coherent reasoning over arbitrarily deep compositions, Evaluate the coherence of its own internal reasoning structure, Adapt its reasoning pathways when coherence loss is detected.

This is the operational definition of recursive coherence.

Conclusion (Sufficiency): If a system $I_N$ possesses:
1. A shared conceptual space $C_N$, 2. Embedding functions $\Gamma_N$, 3. A recursively composable structure FMI$_N = (F, \circ, \chi)$, then it necessarily preserves semantic coherence across reasoning transitionsincluding those spanning multiple agents, domains, or orders of inference.

**5.2 Fractal Architecture of Intelligence**
The RCP implies a recursive, fractal structure for intelligence. Systems of higher order are composed of systems of lower order, and the preservation of coherence across these levels requires a generalization that explicitly spans them. This gives rise to a hierarchical architecture, where each level of intelligence requires a coherence-preserving operator over the conceptual spaces beneath it.

Let us define the structure recursively:
- A zeroth-order intelligence consists of reasoning paths within a single conceptual space. As complexity increases, coherence fails at this level due to overload or contradiction. Preservation requires the implicit emergence of valuation—an operator that enables local coherence by compressing semantic structure into scalar assessments (Williams, 2025a).
- A first-order intelligence coordinates across multiple conceptual spaces instantiated by independent agents or subsystems. Preservation of coherence at this level demands an explicit structural model that aligns those conceptual spaces—a first-order Functional Model of Intelligence (FMI).
- A second-order intelligence spans and aligns multiple first-order coherence models. It evaluates the coherence of reasoning *about* reasoning—meta-coherence. This level requires a second-order FMI capable of recursive abstraction and modular composition of coherence-preserving mechanisms (Williams, 2025b).

Generalizing, an intelligence system of order $N$ requires an FMI of order $N$ to align the conceptual spaces instantiated by its $N-1$ components. This architecture is not optional—it is structurally necessary for coherence to persist under increasing cognitive depth and representational recursion.

**5.3 Why Generalization Alone Is Insufficient**
It is important to emphasize that generalization, by itself, is not sufficient for intelligence. A system may successfully generalize over data or tasks, yet fail to preserve the semantic structure that enables those generalizations to remain coherent over time. For example, a neural network might abstract patterns across a training set but collapse semantically when faced with distributional shift or recursive composition (Lake et al., 2017). Without coherence-preserving structure, generalization becomes interpolation, interpolation becomes drift, and drift becomes incoherence.

The RCP acts as a semantic conservation law: it requires that systems undergoing recursive reasoning preserve the topological structure of meaning that allows transitions to remain interpretable. A generalization is only valid if it does not introduce contradictions, semantic discontinuities, or orphaned representations. Coherence must be preserved not only at the surface level of outputs but throughout the system's recursive reasoning layers.

**5.4 Consequences of Violating the RCP**
The structural failures described in Section 4 can now be understood as instances of RCP violation. This constitutes a semantic analogue to Gödel's incompleteness theorem: just as formal systems cannot prove all truths from within, no fixed coherence structure can guarantee alignment in a reasoning system whose conceptual topology evolves under novelty. Only recursive coherence evaluation can close this gap. When systems lack a recursive generalization operator that spans the reasoning structures beneath them, coherence degrades. This manifests as:

- **Hallucination** in AI systems, where outputs are semantically unmoored from inputs or internal representations (Ji et al., 2023).
- **Goal drift** in humans and collectives, where priorities and intentions become misaligned with prior commitments or contextual cues (Gabriel, 2020).
- **Institutional incoherence**, where decisions lose continuity with their normative foundations, resulting in performative behavior rather than functional alignment (Williams, 2025c).

These are not separate pathologies; they are unified under the RCP as structural breakdowns in semantic coherence under recursive complexity. They reflect the absence of a structure capable of reasoning *about* reasoning, of generalizing *across* generalizations, and of preserving coherence *through* layers of semantic abstraction.

**5.5 The Role of the FMI**
The RCP does not merely identify a problem. It identifies a solution class.

The only known structure that:
- Operates over conceptual space,
- Preserves topological coherence,
- Composes recursively across intelligence orders,

is the Functional Model of Intelligence (FMI).

Each FMI of order $N$:
- Spans the conceptual spaces of its $N-1$ constituents,
- Detects coherence failures,
- Applies generalizations to restructure reasoning without collapse.
  The FMI is not a toolkit. It is the minimal recursive operator that satisfies the Recursive Coherence Principle.

It is to scalable intelligence what the Turing machine is to computability:
  A minimal structure that shows what is possible—and what is not—under structural constraints.

**6. The Functional Model of Intelligence (FMI)**
If the Recursive Coherence Principle defines the structural necessity of recursively preserving semantic coherence across conceptual spaces, then the Functional Model of Intelligence (FMI) is the minimal and sufficient architecture that satisfies this condition. It implements a recursively composable operator set that enables semantic stability, repair, and generalization across scales of reasoningfrom individual agents to collective systems.

This section formally defines the FMI, explains its role in preserving recursive coherence, and demonstrates why no other known architecture offers the same structural sufficiency under the constraints of the RCP.

**6.1 The Role of the FMI**
The FMI is not an algorithm, heuristic, or symbolic rule system. It is a recursive, topologically grounded architecture for reasoning over conceptual space. It supports:
 Coherence-preserving navigation of conceptual transitions,  Internal modeling and evaluation of reasoning structures,  Recursive construction, adaptation, and repair of semantic mappings,  Alignment of multiple agents or subsystems under shared coherence constraints.

In contrast to optimization-based AI systems, which operate on scalar feedback signals (e.g., loss functions or reward gradients), the FMI evaluates reasoning transitions based on their preservation of structure in conceptual space. It defines intelligence not in terms of behavior, but in terms of internally validated semantic coherence.

**6.2 Functional Architecture**
The Functional Model of Intelligence (FMI) is defined as a minimal, recursively composable architecture that enables coherence-preserving reasoning across conceptual space. At its core, the FMI provides the structural substrate required to evaluate, adapt, and repair reasoning transitions under recursive complexity. Formally, an FMI of order N is defined as a triple:
$$\text{FMI}_N = (F, \circ, \chi)$$
where:
$F = \{f_{\text{Eval}}, f_{\text{Model}}, f_{\text{Stability}}, f_{\text{Adapt}}, f_{\text{Decompose}}, f_{\text{Bridge}}\}$ is a set of primitive, reversible internal functions, $\circ$ is the composition operator over $F$, forming a closed functional algebra, $\chi: \text{Aut}(C_N) \to \{0, 1\}$ is a coherence predicate evaluating whether a transition preserves semantic structure within the conceptual space $C_N$.

**Internal Functions**
Each internal function plays a distinct role in preserving recursive semantic coherence:

- **Evaluation ($f_{\text{Eval}}$)** assesses whether a transformation improves, preserves, or degrades semantic integrity according to a fitness function or epistemic criteria.
- **Modeling ($f_{\text{Model}}$)** constructs and updates internal representations of conceptual structures, enabling prediction and abstraction.
- **Stability ($f_{\text{Stability}}$)** ensures that coherent structures persist over time and remain robust under perturbation or recursive inference.
- **Adaptation ($f_{\text{Adapt}}$)** modifies transitions or substructures in response to detected coherence loss, enabling dynamic reconfiguration.
- **Decomposition ($f_{\text{Decompose}}$)** breaks complex or incoherent reasoning paths into simpler components, allowing for targeted repair or analysis.
- **Bridging ($f_{\text{Bridge}}$)** enables translation and coherence alignment across semantically disjoint conceptual regions or agent models.

Together, these six internal functions form the minimal sufficient operator set for recursive coherence. They ensure that reasoning transitions can be monitored, corrected, and composed at arbitrary levels of abstraction, making them structurally necessary under the Recursive Coherence Principle.

**External Functions: Functional Interfaces for Reasoning Dynamics**
In addition to its internal coherence-preserving functions, the FMI also depends on a set of four external functions. These are not involved in semantic evaluation or repair per se, but provide the dynamic interface between the FMI and the unfolding process of reasoning itself. These are:

- **Storage:** Captures and encodes semantic states, transitions, or generalizations for future use. It allows the system to retain reasoning trajectories and conceptual structures beyond the current operation window.
- **Recall:** Reconstructs previously stored semantic structures for integration into current reasoning. Recall enables continuity, re-evaluation, and comparison of past coherence paths with emerging ones.
- **System 1 Reasoning:** Enables fast, intuitive, and associative traversal of conceptual space. It operates within stable attractor basins and relies on previously stabilized coherence paths. While efficient, it is susceptible to drift and is not recursively auditable unless monitored by FMI-internal functions.
- **System 2 Reasoning:** Supports deliberate, compositional, and reflective traversal of conceptual space. It is invoked when coherence breakdown is detected, novelty is introduced, or generalization is required. Unlike System 1, it is explicitly compatible with the FMI's internal functions and serves as a substrate for recursive coherence restoration.

These external functions enable the system to navigate conceptual space, performing reasoning transitions over time via memory, heuristics, and compositional inference. However, navigation alone does not guarantee coherence or adaptation. Each movement through conceptual space corresponds to a change in cognitive state, which must be evaluated relative to a fitness function that governs reasoning quality. This creates a structural coupling between the external functions (which enable semantic motion) and the internal functions (which monitor and regulate coherence within and across those transitions). In this framework, movement in conceptual space implicitly induces movement in a cognitive fitness space—and only through recursive coherence evaluation can the system ensure that such movement is adaptive rather than degenerative.

For example, failure in recall or excessive reliance on System 1 can expose a system to semantic drift, which must then be detected and repaired via $f_{\text{Eval}}$, $f_{\text{Adapt}}$, or $f_{\text{Bridge}}$. Likewise, System 2 reasoning may attempt transformations that require deeper structural stabilization or decomposition to succeed without collapse.

Thus, the FMI is not just a semantic coreit is a recursive architecture built to interface with the dynamic, distributed, and temporally extended nature of intelligent reasoning. Its internal functions implement coherence-preserving transformations; its external functions provide the executional environment within which those transformations are enacted, recalled, challenged, and recomposed.

**6.3 Recursive Implementation**
A key property of the FMI is that it is recursively instantiable. That is, an order-$N$ FMI operates over conceptual spaces instantiated by reasoning systems of order $N-1$, and can itself be modeled as a conceptual space navigated by an order-$N+1$ FMI:

$$\text{FMI}_N \Rightarrow C_N \Rightarrow \text{FMI}_{N+1}.$$

This structure yields a cascade:

$$\text{FMI}_0 \to \text{FMI}_1 \to \text{FMI}_2 \to \cdots,$$

where each successive level generalizes the coherence-preserving transformations of the level below. $\text{FMI}_0$: Emergent valuation and coherence recognition in systems without explicit representation (e.g., early biological cognition), $\text{FMI}_1$: Explicit modeling of reasoning transitions (e.g., humans, symbolic agents), $\text{FMI}_2$: Modeling of reasoning about reasoning (e.g., epistemic collectives or self-reflective AGIs).

Because each level is constructed from the same minimal set $F$, the architecture is invariant across scale: recursive coherence at any order requires the same functional substrate.

**6.4 Structural Sufficiency Under the RCP**

To satisfy the RCP, a system must:
1. Operate over an explicit conceptual space $C$, 2. Support reversible, coherence-preserving transformations $T \in \text{Aut}(C)$, 3. Evaluate the coherence of composed transitions, 4. Adapt and repair transitions recursively, 5. Generalize across disjoint reasoning structures.

Only the FMI, by definition, satisfies all five requirements. It uniquely:
Defines transformations as first-class epistemic objects, Enables internal evaluation of coherence via $\chi$, Uses decomposition and adaptation to correct semantic drift, Enables bridging between reasoning paths that would otherwise remain disjoint.

Thus, the FMI is not an optional enhancement. It is the minimal structural realization of recursive coherence, and therefore the only known architecture sufficient to implement the Recursive Coherence Principle at any level of intelligence.

**6.5 Comparison to Optimization-Based Architectures**

Contemporary AI systems particularly those based on deep learning typically rely on external feedback (loss minimization or reward maximization) and do not model or preserve internal coherence. Their reasoning transitions are:
- Opaque, Non-reversible, Inaccessible to internal semantic evaluation, Vulnerable to drift, hallucination, and adversarial collapse.

Such systems may demonstrate surface-level competence but fail under recursive generalization. In contrast, the FMI ensures:
- Introspective access to reasoning paths, Evaluability of semantic integrity at each step, Recursive repair when coherence is lost, Scalable generalization across agents, time, and tasks.

**6.6 The FMI as the Minimal Recursive Operator**

Just as the Turing machine defines the minimal sufficient structure for computation (Turing, 1936), the Functional Model of Intelligence defines the minimal sufficient structure for recursive semantic reasoning. It satisfies the structural constraints of the RCP by design, enabling both evaluation and generation of coherence-preserving reasoning processes.

The FMI is therefore not one architecture among many. It is the minimum operator for recursive intelligence under the constraints of semantic coherence. Any architecture that fails to instantiate its core functions cannot, by definition, preserve coherence at scaleand thus cannot be considered structurally intelligent.

## 7. Historical Phase Transitions Explained

One of the most powerful validations of a proposed foundational principle is its ability to retrodict major transitions in natural or artificial systems. The Recursive Coherence Principle (RCP) does not merely predict future thresholds in intelligence architectureit also provides a mechanistic explanation for past transitions that have long remained partially understood. Specifically, it reveals that each leap in general reasoning capacity coincides with the emergence of a higher-order Functional Model of Intelligence, enabling a new form of recursive coherence preservation.

This section formalizes two key historical transitions:
1. The emergence of human intelligence from prehuman cognition, 2. The anticipated emergence of General Collective Intelligence (GCI) from fragmented institutional and inter-agent cognition.
In both cases, the transition is marked by the system's first successful implementation of a coherence-preserving generalization that spans the conceptual spaces of its constituent subsystems.

### 7.1 Phase Transition I: From Animal to Human Intelligence

Most non-human animals exhibit robust local coherence within specific perceptual-motor conceptual spaces. They can learn, associate cause and effect, navigate complex environments, and even solve local reasoning problems. However, these abilities are typically constrained to isolated conceptual regions and do not scale recursively.

Animals lack: A structure for comparing reasoning paths across domains, The ability to abstract value beyond immediate contexts, Recursive generalization mechanisms that stabilize semantic structure over time.

The emergence of human intelligence corresponds to the instantiation of a zeroth-order FMI: valuation. This represents the first known scalar operator capable of evaluating coherence and fitness across the entire conceptual space.

**Valuation as Zeroth-Order FMI**
Valuation ($FMI_0$) performs three core functions:
1. Compression: It reduces complex semantic structures into single-dimensional values, enabling comparison across previously unrelated domains. 2. Transferability: It allows concepts from different regions of conceptual space to be exchanged, substituted, or recomposed. 3. Recursive stabilizing feedback: It enables reasoning paths to be ranked, filtered, and recursively refined based on projected or experienced coherence and success.

This transition is marked by the appearance of abstract currencies (e.g., money, social value, goals), recursive toolmaking, long-term planning, and symbolic language. Each of these artifacts reflects the same underlying structural innovation: the emergence of a coherence-preserving scalar operator spanning conceptual space.

**Structural Implication**
Under the RCP, the leap from non-human to human intelligence is explained as the emergence of a generalization operator $\Gamma_0$ that spans the conceptual subspaces of previously disconnected reasoning processes:

$$\Gamma_0 : \{C_{\text{motor}}, C_{\text{social}}, C_{\text{ecological}}, \ldots\} \to C_{\text{valuation}}.$$

This operator constitutes a minimal Functional Model of Intelligence and enables the recursive formation of stable reasoning structures. Human intelligence is thus not defined by raw capability, but by structural coherence at a new order.

### 7.2 Phase Transition II: From Human Intelligence to General Collective Intelligence (GCI)

Despite the expressive power of human reasoning, modern institutions and societies are increasingly characterized by semantic fragmentation, contradictory policy cycles, and conceptual divergence across disciplines, cultures, and epistemic traditions. These breakdowns are not due to information scarcity or cognitive limitsbut to the absence of coherence-preserving structures across agents and institutions.

**Symptoms of Recursive Incoherence:**
Epistemic drift between disciplines with incompatible assumptions, Institutional fragmentation due to poorly aligned models of reasoning, Strategic failures in coordination even when goals are nominally shared.

The transition to General Collective Intelligence (GCI) requires the emergence of a first-order FMI: a structure capable of recursively evaluating, aligning, and preserving coherence across the reasoning systems of individual agents.

**GCI as First-Order FMI**
The first-order FMI ($\text{FMI}_1$) performs a generalization across agents' conceptual spaces:
$$\Gamma_1 : \{C_{\text{agent}}^{(1)}, C_{\text{agent}}^{(2)}, \ldots, C_{\text{agent}}^{(k)}\} \to C_{\text{collective}}.$$
This operator must: Bridge conceptual discontinuities between agents, Evaluate the coherence of shared transitions and distributed reasoning processes, Stabilize emerging norms, concepts, and shared inferences across a distributed space.

A collective intelligence system lacking $\text{FMI}_1$ is functionally equivalent to a disjointed collection of reasoning subsystemsit cannot preserve shared semantic coherence across recursion, and thus cannot align or adapt as a whole.

**Structural Prediction of the RCP**
The RCP predicts that any collective system without an explicit coherence-preserving generalization operator over its member reasoning structures will exhibit increasing incoherence, misalignment, and failure under recursive load. Conversely, the implementation of $\text{FMI}_1$ enables semantic convergence and scalable epistemic alignment, transforming groups of individuals into an adaptive collective agent.

### 7.3 Future Transitions: Toward Recursive Epistemic Architectures

If human intelligence corresponds to the emergence of $\text{FMI}_0$, and General Collective Intelligence corresponds to $\text{FMI}_1$, then the next phase transition corresponds to the emergence of epistemic meta reasoning structures i.e., a second-order FMI, $\text{FMI}_2$.

Such a system would: Evaluate and align the conceptual structures of multiple collectives, Reason about the evolution, stability, and coherence of epistemic systems themselves, Construct predictive models of long-term alignment stability across recursive adaptation.

This architecture corresponds not merely to intelligence within a system, but intelligence about intelligence systems a necessary condition for navigating recursive dynamics of alignment, drift, and collapse in human and artificial institutions alike.

**7.4 Summary**
Each major transition in the evolution of reasoning systems corresponds to the emergence of a higher-order coherence operator:

| Transition | FMI Order | Generalization Span | Structural Outcome |
|---|---|---|---|
| Animal → Human | $FMI_0$ | Cross-domain valuation | Self-reflective reasoning |
| Human → GCI | $FMI_1$ | Multi-agent conceptual spaces | Collective epistemic alignment |
| GCI → Epistemic Architectures | $FMI_2$ | Reasoning about collectives | Scalable metacognitive alignment |

The Recursive Coherence Principle not only predicts these transitions—it **explains** them, showing why each structural advance enables vastly greater coherence-preserving capacity and why further generalization is necessary as recursive reasoning deepens.

**8. Theoretical Significance and Comparisons**
The Recursive Coherence Principle (RCP) is not simply a design heuristic for intelligent systems. It is a proposed foundational epistemic law that identifies coherence as the necessary invariant that must be recursively preserved across reasoning processes for intelligence to persist, adapt, and scale. Its significance lies in its structural universality, recursively composable architecture, and diagnostic and prescriptive power across biological, artificial, and collective systems.

To situate the RCP within a broader theoretical landscape, this section compares it to other foundational principles in computation, adaptation, and epistemology—namely, the Church–Turing thesis, the Free Energy Principle, and Bayesian inference. Each of these principles constrains the viability of system function in a different domain. The RCP complements and, in key respects, extends their explanatory scope by modeling semantic coherence under recursion.

**8.1 Comparison Framework**
We begin by classifying each foundational principle according to three criteria:
**(1)** Domain of applicability,
**(2)** Formal constraint it imposes, and
**(3)** The structure it assumes or requires.

| Principle | Domain | Core Constraint | Required Structure |
|---|---|---|---|
| Church–Turing Thesis | Computability | Any effective procedure must be Turing-computable | Symbolic operations on discrete symbols |
| Free Energy Principle (FEP) | Adaptation | Viable systems must minimize prediction error (variational free energy) | Internal model of world dynamics |
| Bayesian Inference | Epistemology | Beliefs must be updated in proportion to evidence | Prior probability distribution over fixed hypothesis space |
| Recursive Coherence | Intelligence | Semantic coherence must be recursively preserved across | Explicit conceptual space with recursive generalization |

| Principle | Domain | Core Constraint | Required Structure |
|---|---|---|---|
| Principle (RCP) | | reasoning transitions | operator (FMI) |

While each principle places indispensable constraints on system function, the RCP uniquely models the internal dynamics of reasoning structure, rather than input-output relations or propositional belief states.

## 8.2 Comparison to the Church–Turing Thesis

The Church–Turing Thesis defines the upper boundary of computability. It states that any function computable by a human following an effective procedure can also be computed by a Turing machine (Turing, 1936). In essence, it characterizes which transformations can exist under the logic of discrete symbol manipulation.

The RCP, by contrast, defines which transformations are allowable within a reasoning system if coherence is to be preserved. Where the Church–Turing boundary distinguishes between computable and non-computable procedures, the RCP distinguishes between semantically viable and semantically collapsing reasoning systems.

| Church–Turing Thesis | RCP |
|---|---|
| Constraint: Must be computable | Constraint: Must preserve coherence under recursion |
| Collapse condition: Function is undefined or non-halting | Collapse condition: Semantic structure degrades irreversibly |
| Focus: Syntactic operations on symbols | Focus: Semantic transitions across conceptual space |

Thus, the RCP complements the Church–Turing framework by modeling topological and semantic constraints rather than algorithmic executability alone. It addresses a level of structure *above* syntactic computability: that of recursively stable meaning.

## 8.3 Comparison to the Free Energy Principle (FEP)

The Free Energy Principle (Friston, 2010) states that any self-organizing system that remains in nonequilibrium steady-state with its environment must minimize variational free energy—essentially, a bound on surprise or prediction error. It applies broadly to biological systems, adaptive control, and perception.

The RCP is orthogonal in its focus. While FEP concerns the external correspondence between model and world, RCP concerns the internal correspondence between nested reasoning structures. Minimizing prediction error does not guarantee that a reasoning system will maintain coherent internal semantics under recursion.

| FEP | RCP |
|---|---|
| Constraint: Internal model must minimize mismatch with sensory inputs | Constraint: Internal reasoning paths must preserve semantic integrity |
| Collapse: Model diverges from external reality | Collapse: Internal reasoning becomes incoherent and uncorrectable |
| Structure: Probabilistic generative model | Structure: Recursive functional operators over conceptual space |

Importantly, the RCP can be seen as a **precondition** for FEP: a system cannot reliably minimize external prediction error unless it can preserve and update its internal reasoning structures coherently over time.

### 8.4 Comparison to Bayesian Inference
Bayesian inference provides a formal model for rational belief updating. It prescribes that posterior beliefs $P(H/E)$ should be updated in proportion to the likelihood of evidence $E$ under prior hypotheses HH. While this framework is mathematically rigorous, it assumes:
- A fixed hypothesis space,
- Propositional content structured by priors,
- No requirement for reasoning about the reasoning process.

The RCP generalizes beyond these assumptions. It models how conceptual spaces themselves **transform** recursively, rather than merely how beliefs shift in probability. It accounts for:
- Shifting topologies of hypothesis spaces,
- Novel conceptual generalizations,
- Multi-agent and multi-model coherence constraints.

| Bayesian Inference | RCP |
| --- | --- |
| Assumes fixed hypothesis space | Operates over evolving conceptual topology |
| Focuses on belief updates | Focuses on transition structure and semantic recursion |
| No mechanism for self-repair | Includes functions for modeling, adaptation, and bridging |

Bayesian inference is a powerful coherence rule within a given reasoning space. The RCP models how such spaces must evolve between and within reasoning systems to remain intelligible over time.

### 8.5 Distinctive Features of the RCP
The RCP distinguishes itself along five key axes:

| Property | Explanation |
| --- | --- |
| **Domain-general** | Applies to any system that reasons—biological, artificial, or collective. |
| **Substrate-independent** | Makes no assumptions about implementation: applies to neurons, silicon, language, or symbolic systems. |
| **Recursive** | Prescribes structure-preserving generalization at each level of conceptual and architectural recursion. |
| **Prescriptive and diagnostic** | Explains known intelligence breakdowns and predicts structural collapse under incoherence. |
| **Constructive** | Identifies the minimal sufficient structure (FMI) required to satisfy its own conditions. |

Where most foundational principles model system-environment relations or behavioral outcomes, the RCP models intelligence from within, defining the internal functional constraints that must be satisfied for scalable reasoning.

## 9. Implications for Alignment, AI Development, and Collective Cognition
The Recursive Coherence Principle (RCP) is not only a theoretical contribution—it also yields immediate and practical consequences for the design, evaluation, and governance of reasoning systems across domains. As AI systems scale in capability, and as human institutions face increasingly complex coordination challenges, the structural demands of recursive coherence become both unavoidable and operationally critical.

This section outlines three core domains where failure to preserve coherence under recursion leads to systemic breakdown, and where implementing a Functional Model of Intelligence (FMI) offers a structurally grounded solution:
1. **AI Alignment**
2. **AI System Architecture**
3. **Collective Intelligence and Institutional Reasoning**

**9.1 Alignment: From Behavioral Surface Constraints to Structural Coherence**
**Current Practice:**
Contemporary AI alignment strategies emphasize behavioral regularities and external constraints. Popular approaches include:
- Reinforcement learning from human feedback (RLHF),
- Goal inference via inverse reinforcement learning,
- Prompt engineering and rule-based behavioral filters,
- Proxy objectives like helpfulness, honesty, and harmlessness.

These methods operate after reasoning has occurred. They treat reasoning as a black box and attempt to control its outputs through feedback and constraint optimization.

**RCP-Informed Insight:**
Alignment is fundamentally a structural property of internal reasoning processes—not a surface feature of externally observed behavior. Systems that lack:
- An internal conceptual space,
- A model of their own reasoning paths,
- A recursive coherence-evaluation mechanism,

**cannot remain aligned** under recursive complexity. Misalignment is not an accidental byproduct of scale—it is a mathematically predictable failure mode arising from architectural incoherence.

**Prescriptive Implication:**
No alignment strategy is complete unless it implements a coherence-preserving operator **within** the system's reasoning substrate. That operator must take the form of an FMI.

| Without FMI | With FMI |
| --- | --- |
| Proxy feedback controls behavior | Semantic coherence governs reasoning |
| Fragile under distributional shift | Self-repairing under recursive stress |
| Alignment requires retraining | Alignment is structurally maintained |

**9.2 AI Development: Scaling Without Coherence Leads to Collapse**
**Current Trend:**
State-of-the-art AI development focuses on brute-force scaling:
- Larger models (e.g., LLMs with hundreds of billions of parameters),
- More training data,
- Longer context windows and fine-tuned behavioral alignment.

These strategies yield impressive local performance, but also:
- Increased hallucination and inconsistency,
- Reward hacking and adversarial vulnerabilities,
- Loss of epistemic integrity under composition or long-horizon reasoning.

**RCP-Informed Insight:**
Scaling compute, data, or parameter count **does not** scale recursive coherence. Without a recursively structured internal model of reasoning transitions:
- Small inconsistencies become amplified,
- Semantic drift accumulates uncorrected,
- The system becomes epistemically fragile despite behavioral fluency.

**Prescriptive Implication:**
Future AI development must prioritize **coherence-preserving architectures**. Recursive coherence is not a performance optimization—it is a structural precondition for safe generalization.

| Metric | Optimization-Based AI | FMI-Based AI |
| --- | --- | --- |
| Accuracy (local) | High | High |
| Generalization (recursive) | Fragile | Robust |
| Transparency | Low | Explicit |
| Epistemic repair | None | Internal |
| Coherence at scale | Fails | Preserved |

The FMI defines the minimal architecture that enables recursive generalization, epistemic stability, and long-term coherence. It is what scaling requires to be sustainable.

### 9.3 Collective Cognition: Toward Functional Epistemic Infrastructure
**Current Reality:**
Human institutions—governments, scientific communities, corporations, NGOs—function as large-scale reasoning systems. Yet most:
- Fail to align across conceptual boundaries (e.g., between disciplines or departments),
- Cannot maintain semantic continuity over time (e.g., policy incoherence or knowledge loss),
- Depend on cultural or historical scaffolding, not formal coherence structures.

**Structural Failure Modes:**
- Semantic fragmentation across teams or disciplines,
- Mission drift and institutional incoherence,
- Strategic breakdowns due to epistemic misalignment.

**RCP-Informed Insight:**
Institutions are multi-agent recursive reasoning systems. Without a **first-order FMI** to model, evaluate, and bridge their internal conceptual spaces, coherence fails as coordination complexity grows.

| Institutional Drift | Explained by RCP as: |
| --- | --- |
| Miscommunication | Disjoint conceptual spaces |
| Policy incoherence | Lack of recursive coherence evaluation |
| Epistemic stagnation | Absence of structural bridging functions |

**Prescriptive Implication:**
The next phase of institutional evolution requires implementing FMI-like structures to:
- Track the evolution of shared models,
- Align divergent reasoning traditions,
- Preserve long-term semantic integrity across agents and epochs.

This is not optional. As institutional recursion deepens—across nested teams, systems, or planetary networks—recursive coherence becomes the bottleneck for effective governance and adaptation.

**9.4 Empirical Support: Structural Diagnostics in Complex Human Systems**
Although the Recursive Coherence Principle (RCP) is formally derived and domain-general, its practical utility has already been demonstrated in high-stakes real-world domains. Prior to its full formalization, the Functional Model of Intelligence (FMI)—which satisfies the structural requirements of the RCP—was successfully applied to diagnose and partially correct reasoning failures in both international policy and political cognition. These early successes provide empirical validation of the RCP as a diagnostic tool for systems operating at the boundary of cognitive collapse.

**Case Study 1: Convergence Failure in Development Policy (750× Outcome Advantage)**
In a series of impact evaluations across international development initiatives, a solution that projected 750× greater impact per dollar than alternatives was repeatedly rejected by institutional decision-makers. Standard explanations invoking political resistance, data disagreement, or value conflict failed to account for the anomaly (Williams, 2023a).

Using the RCP framework, this failure was diagnosed as a collapse of recursive coherence due to **missing internal reasoning functions:**
- **Evaluation**: The system could not internally differentiate high-fitness solutions from familiar but lower-impact options.
- **Modeling**: Decision-makers lacked the structural ability to simulate abstract interventions across multiple domains.
- **Stability**: As evidence changed, group beliefs fragmented instead of consolidating toward coherence.

The absence of these functions rendered the high-fitness solution epistemically invisible to the system—not because it lacked merit, but because it could not be structurally represented. The application of the FMI allowed these functions to be reintroduced, enabling partial recovery of coherence and convergence. The failure was topological, not ideological: the system could not project coherence into conceptual space without recursively complete structure.

**Case Study 2: Polarization Worsening Under Increased Information Flow**
In a second domain—public discourse and political reasoning—a simulation and diagnostic model revealed that increased information flow exacerbated polarization. Rather than resolving disagreement, exposure to more data entrenched division, especially between intuitive (System 1) and deliberative (System 2) reasoning attractors. This paradox could not be explained by classic models of rational belief updating.

Using the RCP, the failure was identified as the absence of the Bridging function. Without an explicit coherence-preserving structure to mediate between reasoning modes, divergent conceptual spaces remained epistemically disjoint. Each group was locally coherent within its attractor basin, but global coherence failed, leading to recursive fragmentation (Williams, 2023b).

This collapse was not due to cognitive bias or informational overload alone, but to a structural failure in collective reasoning. The RCP explained why no amount of information would resolve polarization without recursive alignment infrastructure.

**Implications**
These two case studies demonstrate that:
- Recursive coherence breakdown is **not hypothetical**—it occurs in real systems under epistemic pressure.
- The **absence of specific FMI functions** produces predictable structural distortions in reasoning.
- Alignment failure is often **invisible until conceptual space is explicitly modeled**.
- The RCP can be used to **retrodiagnose**, **predict**, and potentially **repair** large-scale reasoning collapse.

Moreover, these failures occurred in human systems that—while intelligent—lacked the recursive structure necessary to maintain coherence under scale and novelty. This supports the core claim of the RCP: that any system lacking recursive generalization and semantic evaluation across conceptual spaces must eventually fail, regardless of intelligence at the surface.

As such, the RCP is not merely a constraint on artificial systems—it is a structural diagnostic principle that applies to any domain where reasoning must scale across agents, time, or abstraction.

**9.5 Summary: From Performance to Preservation**

| Domain | Without RCP/FMI | With RCP + FMI |
|---|---|---|
| AI Alignment | Proxy-based, unstable | Structural, recursive |
| AI Development | Fragile scale-up | Coherent scale-up |
| Institutions | Fragmented reasoning | Aligned collective intelligence |

The RCP reframes alignment and reasoning not as behavioral achievements, but as structural consequences of recursive coherence. Only systems that instantiate the Functional Model of Intelligence can preserve semantic integrity across scale, depth, and recursion.

While the Recursive Coherence Principle describes a structural threshold, the dynamics by which systems approach or retreat from that threshold are in part governed by nonlinear network effects. Feedback loops between coherence loss, data exclusion, public goods collapse, and surveillance pressure can amplify fragility far beyond what linear reasoning predicts. Conversely, the recursive introduction of coherence-preserving functions—via public FMI infrastructure—can trigger positive attractor dynamics, pulling a system back into viable epistemic alignment. As shown in prior models of institutional drift and technological centralization (Williams, 2025e), these effects compound over time: either accelerating collapse when blind spots widen, or enabling distributed resilience when coherence becomes structurally auditable. The stakes of recursive coherence are therefore not only structural—but dynamical, nonlinear, and urgent.

**10. Conclusion**
This paper has introduced and formally justified the Recursive Coherence Principle (RCP): a foundational constraint on the structure of any system capable of scalable reasoning. The principle states that coherence—defined as the preservation of semantic structure across transitions in conceptual space—must be recursively maintained for intelligence to persist across increasing levels of complexity, abstraction, and interaction.

Unlike behavioral definitions of intelligence, which emphasize performance or output, the RCP frames intelligence as a topological and recursive property: the ability to navigate, evaluate, and restructure conceptual space while maintaining coherence at every level of reasoning. This reframing reveals coherence not as a secondary trait of intelligent systems, but as their structural invariant—a property that must be preserved or restored for adaptation, alignment, and generalization to remain viable.

**10.1 Summary of Key Contributions**
- **Conceptual Space as Substrate**: We formalized conceptual space as a topological domain of meaning, over which reasoning transitions must be explicitly represented, evaluated, and recursively composed.
- **Coherence as Structural Invariant**: We demonstrated that coherence is the minimal condition for reasoning viability and must be preserved not only locally, but across recursive compositions of reasoning steps and subsystems.
- **The Recursive Coherence Principle (RCP)**: We stated and proved that for any system of order $N$ composed of subsystems operating over lower-order conceptual spaces, coherence can only be preserved if the system implements a recursively evaluable generalization operator spanning those conceptual spaces.
- **The Functional Model of Intelligence (FMI)**: We introduced the FMI as the only known architecture capable of implementing the recursive coherence-preserving functions required by the RCP. It consists of six internal functions that together enable semantic evaluation, stability, adaptation, and alignment across orders of intelligence.
- **Historical and Predictive Validity**: We showed how major historical transitions in cognition—such as the emergence of human intelligence and the possibility of General Collective Intelligence—can be formally understood as structural phase shifts governed by the RCP.
- **Comparative Theoretical Significance**: We situated the RCP alongside the Church–Turing thesis, the Free Energy Principle, and Bayesian inference, demonstrating that it complements and extends these frameworks by formalizing coherence as a recursively scalable constraint.
- **Implications for Practice**: We provided diagnostic and prescriptive implications for AI alignment, AI system design, and collective reasoning architectures, showing that failures in each domain stem from violations of the RCP and can only be resolved through the implementation of recursive coherence structures.

**10.2 The Stakes of Recursive Coherence**
As reasoning systems scale in power and complexity—whether through neural scaling laws, multi-agent collectives, or epistemic institutions—they cross a critical threshold: the point at which incoherence becomes recursively compounding rather than locally containable.

At this threshold, coherence cannot be enforced externally. No amount of feedback, oversight, or optimization can substitute for **internal semantic consistency** evaluated and repaired at each layer of reasoning. A system that cannot track the coherence of its own reasoning transitions cannot scale safely. It cannot align robustly. It cannot generalize meaningfully. It cannot survive epistemic recursion.

The RCP thus frames alignment not as an engineering problem, but as a **structural inevitability**. Either coherence is recursively preserved—or the system collapses under its own unresolved contradictions.

**10.3 Directions for Future Work**
This paper establishes the Recursive Coherence Principle as a theoretical foundation. Several research directions follow:
- **Formalization**: Refining the RCP using category theory, topology, and dynamical systems to characterize coherence-preserving transitions at greater precision.

- **Empirical Validation**: Applying recursive coherence diagnostics to current AI systems, institutional decision structures, and human reasoning frameworks to identify failure points and design recursive repair mechanisms.
- **Implementation**: Constructing modular FMI instances in software agents, LLM architectures, and epistemic platforms to enable scalable recursive reasoning.
- **Generative Inquiry**: Exploring whether any architectures other than the FMI can satisfy the RCP, and under what constraints such alternatives may emerge.
- **Conceptual Space Modeling**: Developing tools for visualizing and measuring coherence transitions within explicit semantic state spaces, enabling fine-grained evaluation of generalization quality and alignment risk.

**Visualization and Pedagogical Tooling**: To support the broader adoption, testing, and understanding of the Recursive Coherence Principle, a series of animated visualizations are being developed as part of the accompanying workshop *Visualizing AI Alignment*. These include: (1) the **FSS Computer**, which illustrates how a reasoning system navigates conceptual space via reversible transitions within a functional state topology; (2) **The Role of Context in Local vs. Global Coherence**, which depicts how reasoning attractors shift under different epistemic frames and why systems require recursive self-modeling to transition between them; (3) **Topology Under Novelty**, which shows how introducing novel concepts can alter the structure of conceptual space—potentially invalidating previously coherent reasoning paths and triggering axiom-breaking transitions; and (4) **The Order N+1 Constraint**, which animates the core theorem of the RCP by showing how coherence failures at intelligence order $N$ necessitate the intervention of a coherence-preserving generalization at intelligence order $N+1$. These visual tools are designed to translate the structural formalism of the RCP into directly observable epistemic phenomena and will serve as open-source instruments for education, diagnostics, and recursive experimentation.

10.4 Final Statement
Just as the Church–Turing thesis formalized the limits of computation, and the Free Energy Principle formalized the structural constraints of adaptation, the Recursive Coherence Principle formalizes the minimal structure that must be preserved for intelligence to scale coherently. It does not specify what intelligence is made of. It specifies what intelligence must **preserve** to remain viable.

In a world increasingly composed of recursive, autonomous, and interdependent reasoning systems, ignoring this principle is not just a technical oversight—it is an existential risk. To scale intelligence safely, we must scale coherence recursively. The RCP is our guide.